\documentclass[11pt]{article}

\usepackage[utf8]{inputenc}
\usepackage[T1]{fontenc}
\usepackage{lmodern}
\usepackage{microtype}
\usepackage{geometry}
\usepackage{float}
\usepackage{amsmath,amssymb,mathtools,amsthm,bm}
\usepackage[table]{xcolor}
\definecolor{tablehead}{RGB}{255,244,214}
\definecolor{tablestripe}{RGB}{255,251,240}

\definecolor{tabgreenA}{RGB}{232,252,238}
\definecolor{tabgreenB}{RGB}{244,255,248}
\definecolor{headerblue}{RGB}{230,240,255}
\definecolor{rowgray}{RGB}{240,240,240}
\definecolor{cellblue}{RGB}{236,244,255}
\definecolor{cellyellow}{RGB}{255,236,214}
\definecolor{cellgreen}{RGB}{232,248,236}
\usepackage{booktabs}
\usepackage{makecell}
\usepackage{marvosym}
\usepackage{graphicx}
\let\OriginalIncludeGraphics\includegraphics
\renewcommand{\includegraphics}[2][]{%
  \IfFileExists{#2}{%
    \OriginalIncludeGraphics[#1]{#2}%
  }{%
    \fbox{\parbox{0.8\linewidth}{\centering Missing figure: \texttt{\detokenize{#2}}}}%
  }%
}
\usepackage{enumitem}
\usepackage[numbers,sort&compress]{natbib}
\usepackage[colorlinks=true,linkcolor=blue,citecolor=blue,urlcolor=blue]{hyperref}
\usepackage[nameinlink,capitalise]{cleveref}
\usepackage[most]{tcolorbox}
\newcommand{\E}{\mathbb{E}}

\newcommand{\KL}{\mathrm{KL}}
\newcommand{\score}{\mathbf{s}}

\newtcolorbox{questionbox}{
  colback=green!6,
  colframe=green!35!black,
  boxrule=0.6pt,
  arc=1.5mm,
  left=1.2mm,
  right=1.2mm,
  top=1mm,
  bottom=1mm
}

\title{Rethinking the Diffusion Model from a Langevin Perspective}
\author{
  Candi Zheng, Yuan Lan  \\
  Department of Mathematics, The Hong Kong University of Science and Technology
}
\date{}

\begin{document}

% \begin{titlepage}
%   \centering
%   \vspace*{0.16\textheight}

%   {\LARGE\bfseries Rethinking the Diffusion Model from a Langevin Perspective\textsuperscript{*}\par}
%   \vspace{1.4cm}

%   {\large Candi Zheng \quad Yuan Lan\par}
%   \vspace{0.45cm}
%   {\normalsize Department of Mathematics, The Hong Kong University of Science and Technology\par}

%   \vfill
%   {\normalsize \today\par}
% \end{titlepage}

\begin{center}
{\LARGE\bfseries Rethinking the Diffusion Model from a Langevin Perspective\textsuperscript{*}\par}
\vspace{0.9cm}
{\large Candi Zheng\textsuperscript{\Letter} \quad Yuan Lan\par}
\vspace{0.35cm}
{\normalsize Department of Mathematics, The Hong Kong University of Science and Technology\par}
\end{center}
\begingroup
\renewcommand{\thefootnote}{\Letter}
\footnotetext{\texttt{maczheng@ust.hk}}
\addtocounter{footnote}{-1}
\endgroup

\begin{abstract}
Diffusion models are often introduced from multiple perspectives, such as VAEs, score matching, or flow matching, accompanied by dense and technically demanding mathematics that can be difficult for beginners to grasp. One classic question is: how does the reverse process invert the forward process to generate data from pure noise? This article systematically organizes the diffusion model from a fresh Langevin perspective, offering a simpler, clearer, and more intuitive answer. We also address the following questions: how can ODE-based and SDE-based diffusion models be unified under a single framework? Why are diffusion models theoretically superior to ordinary VAEs? Why is flow matching not fundamentally simpler than denoising or score matching, but equivalent under maximum-likelihood? We demonstrate that the Langevin perspective offers clear and straightforward answers to these questions, bridging existing interpretations of diffusion models, showing how different formulations can be converted into one another within a common framework, and offering pedagogical value for both learners and experienced researchers seeking deeper intuition.
\end{abstract}
\begingroup
\renewcommand{\thefootnote}{\fnsymbol{footnote}}
\footnotetext[1]{This article is adapted and organized from the ICLR Blogpost Track post: \url{https://iclr-blogposts.github.io/2026/blog/2026/rethinking-diffusion-langevin/}.}
\endgroup
\tableofcontents
\clearpage

\section{Introduction}

Modern diffusion models are built upon two fundamental processes: the forward process, which gradually corrupts data with noise during training, and the reverse process, which generates data by sampling from noise. The development of diffusion models has diverged into several valuable perspectives, illuminating different aspects of these processes. Most interpretations fall into three main frameworks: the variational autoencoder (VAE) perspective, the score-based perspective, and the flow-based perspective. Although there are many tutorials available, learning the core theory of diffusion models remains challenging for beginners due to mathematically dense derivations and fragmented intuitions scattered across these different perspectives.

The \textbf{VAE perspective} treats the forward diffusion process as an encoder that adds noise to the data and the reverse process as a decoder that removes noise, with the Evidence Lower Bound (ELBO) serving as the training objective \citep{Luo2022UnderstandingDM,Ho2020DenoisingDP}. This framework is straightforward for those familiar with VAEs. However, it is not obvious why the iterative denoising in diffusion models outperforms the one-step decoding typical of ordinary VAEs.

The \textbf{score-based perspective} \citep{Song2020ScoreBasedGM} places a clearer emphasis on the paired relationship between the forward and reverse processes, which contributes to the superiority of diffusion models. It typically introduces the forward process first, then directly presents the reverse process by reverse-time diffusion \citep{Anderson1982ReversetimeDE} without derivation. Understanding the derivation of the reverse process usually requires familiarity with advanced mathematical concepts such as the Kolmogorov backward equations, which makes it less accessible. Additionally, the score matching objective is specifically tailored for score models, making it less straightforward to generalize to other approaches such as flow matching models.

A third valuable viewpoint is the \textbf{flow-based perspective} \citep{liu2022flow}, which has rapidly gained popularity in modern diffusion models. Although this approach is theoretically equivalent to both the VAE and score-based frameworks \citep{gao2025diffusion}, it distinguishes itself by highlighting a clear and intuitive straight-line interpolation between data and noise. This conceptual clarity makes the flow-based perspective accessible and attractive. However, this apparent simplicity can be misleading: it can create the impression that flow matching is fundamentally simpler than denoising or score matching, rather than a mathematically equivalent reformulation.

In this article, we systematically organize the theory of diffusion models and present a perspective that is both mathematically simple and intuitively clear: the \textbf{Langevin perspective}. This approach, relying only on basic techniques from stochastic differential equations (SDEs), provides a straightforward derivation of the reverse process and explains why flow matching is not fundamentally simpler than denoising or score matching, but is equivalent to them under maximum likelihood.

\begin{figure}[H]
  \centering
  \label{fig:central_insight}
  \includegraphics[width=0.9\linewidth]{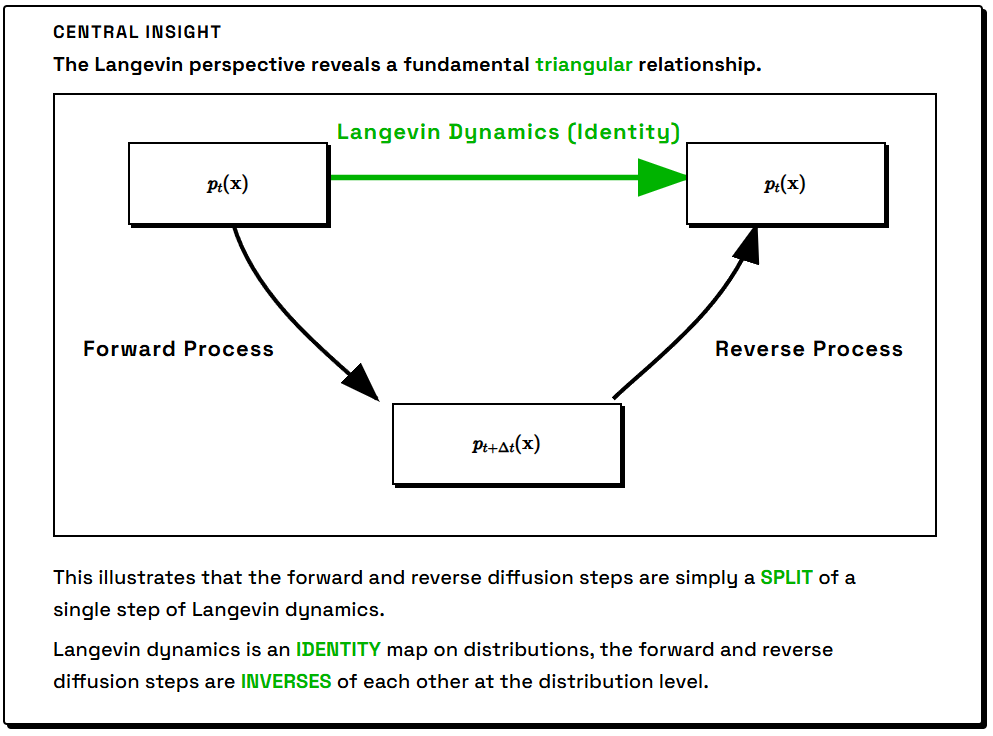}
\end{figure}

% \begin{questionbox}
% \textbf{Guiding questions.}
% \begin{enumerate}[leftmargin=1.5em]
%   \item How can ODE-based and SDE-based diffusion models be unified under a single framework?
%   \item Why are diffusion models theoretically superior to ordinary VAEs?
%   \item Why is flow matching not fundamentally simpler than denoising or score matching, but equivalent under maximum likelihood?
% \end{enumerate}
% \end{questionbox}

% \paragraph{Prerequisites.}
% \begin{itemize}[leftmargin=1.5em]
%   \item Basic calculus.
%   \item Basic probability theory.
%   \item No prior knowledge of stochastic processes is required.
% \end{itemize}

% \begin{tcolorbox}[colback=blue!3,colframe=blue!35!black,title=Conversion note]
% This LaTeX manuscript is a full-content conversion of the original long-form blog post. Interactive HTML widgets, embedded scripts, and custom web layouts have been replaced by static figures, standard tables, and appendix material suitable for arXiv.
% \end{tcolorbox}

\section{Langevin Dynamics as 'Identity' Operation}

This section will show that Langevin dynamics acts as an `identity' operation on distributions, mapping a sample from a distribution to another sample from the same distribution.

Langevin dynamics \citep{Langevin1908} is a stochastic process for sampling from a target probability distribution $p(\mathbf{x})$. One common form is the SDE
\begin{equation}
d\mathbf{x}_t = g(t)\,\score(\mathbf{x}_t)\,dt + \sqrt{2g(t)}\,d\mathbf{W}_t,
\label{eq:langevin}
\end{equation}
where $\score(\mathbf{x}) = \nabla_{\mathbf{x}}\log p(\mathbf{x})$.

At first sight, the extra term $d\mathbf{W}_t$ may make this SDE look much more complicated than an ordinary differential equation (ODE). In fact, it is best to think of it as an ODE with an additional infinitesimal random perturbation at each step. Informally, one can write
\begin{equation}
d\mathbf{W}_t = \sqrt{dt}\,\boldsymbol{\epsilon},
\end{equation}
where $\boldsymbol{\epsilon}$ is a standard Gaussian random noise. The remaining terms are familiar: $\score(\mathbf{x})$ is the score function of $p(\mathbf{x})$, and $g(t)$ is an arbitrary positive function rescaling the time $t$.

This dynamics is often used as a Monte Carlo sampler to draw samples from $p(\mathbf{x})$, since $p(\mathbf{x})$ is its \emph{stationary distribution}---the distribution that $\mathbf{x}_t$ converges to and remains at as $t\to\infty$, regardless of the initial distribution of $\mathbf{x}_0$. For an intuitive derivation of this statement, see \cref{sec:appendix-stationary}.

Langevin dynamics, while widely used for sampling from complex distributions, becomes inefficient in high-dimensional or multimodal settings due to slow mixing and sensitivity to hyperparameters such as step size and noise scale. Nevertheless, it plays a crucial foundational role in diffusion models because of the following property:

\begin{quote}
For a target distribution $p(\mathbf{x})$, Langevin dynamics acts as an identity operation on the distribution, transforming a sample from $p(\mathbf{x})$ into a new, independent sample from the same distribution.
\end{quote}

This ``identity on distribution'' view is the key bridge to diffusion models. Forward and reverse processes can be interpreted as a split of this identity into a noising phase and a denoising phase.

\begin{figure}[H]
  \centering
  \includegraphics[width=0.6\linewidth]{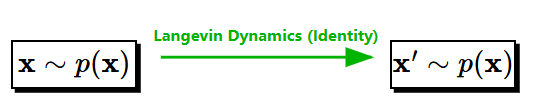}
  \caption{Langevin dynamics acts as an identity operation on $p(\mathbf{x})$: starting from a sample $\mathbf{x} \sim p(\mathbf{x})$, it produces a new sample $\mathbf{x}'$ from the same distribution.}
  \label{fig:langevin-identity}
\end{figure}

The identity viewpoint in \cref{fig:langevin-identity} will be the organizing principle for the rest of this article.

\section{Spliting the Identity into Forward and Reverse Processes}

One key reason Langevin dynamics struggles in high-dimensional settings is the challenge of initialization \citep{song2019generative}. The score function required by it is learned from real data and is therefore reliable only near true data points, while being poorly estimated elsewhere. Yet in generative modeling we need to start from locations that may be far from the data manifold. Finding an initialization that is both realistic and close enough to the true data manifold is difficult, making effective generation with Langevin dynamics challenging in practice. In short, Langevin dynamics is well-suited for generating new samples from an existing one, but ill-suited for generating samples entirely from scratch.

An enhancement to Langevin dynamics is the Annealed Langevin dynamics \citep{song2019generative}. Instead of using a single Langevin sampler, this method involves training a sequence of Langevin dynamics, each corresponding to a different level of noise added to the data. Starting from pure noise, the method gradually reduces the noise level, switching between these samplers at each step. In this way, samples are progressively transformed from random noise into data-like samples, using Langevin dynamics that are effective for each stage of noise contamination. This approach highlights the importance of using multiple noise levels.

Diffusion models take this concept a step further by completely separating the training and inference processes: one process trains the model at different noise levels, while another process samples from noise to generate data. In this section, we show that the forward and reverse processes in diffusion models are splits of a single Langevin dynamics, decomposing the identity operation into a noising phase and a denoising phase.

\subsection{The Forward Diffusion Process for Noising}

The forward diffusion process in diffusion model generates the necessary training data: clean images and their progressively noised counterparts. In continuous time, a very general way to describe such a process is by an It\^o SDE of the form
\begin{equation}
d\mathbf{x}_t = f(\mathbf{x}_t,t)\,dt + g(t)\,d\mathbf{W}_t,\qquad t\in[0,T].
\label{eq:forward-sde}
\end{equation}

where $t \in [0,T]$ is the forward diffusion time, $\mathbf{x}_t$ is the noise-contaminated image at time $t$, $\mathbf{W}_t$ is a Brownian motion, $f(\mathbf{x}_t, t)$ is the drift, and $g(t)$ scales the injected noise. Different choices of $f$ and $g$ correspond to different forward-diffusion parameterizations used in diffusion models.

In practice, diffusion models are usually instantiated by choosing specific parameterizations of this SDE. The most common ones are the \textbf{variance-preserving (VP)} process, implemented in DDPMs as an Ornstein--Uhlenbeck dynamics that gently pulls samples toward the origin while injecting noise so that the marginal converges to a standard Gaussian; the \textbf{variance-exploding (VE)} process, where there is no restoring drift and the noise scale grows with time so that the variance ``explodes''; and \textbf{flow-matching} formulations, which view generation as following a time-dependent flow that implements a ``straight line'' interpolation between data and noise under a carefully designed schedule.

 \cref{tab:forward-processes}  summarizes these three forward processes of different model types, as well as their corresponding SDEs expressed in terms of their respective noise-levels. In what follows, we adopt Karras' notation for the VE parameterization \citep{Karras2022Elucidating}.

\begin{table}[t]
\centering
\caption{Forward processes across model parameterizations.}
\label{tab:forward-processes}
\begingroup
\renewcommand{\arraystretch}{1.5}
\setlength{\tabcolsep}{8pt}
\resizebox{\textwidth}{!}{%
\begin{tabular}{llll}
\toprule
% \rowcolor{tabgreenA}
\textbf{Model Type} & \textbf{Noise-level parameter} & \textbf{Forward process} & \textbf{Forward SDE} \\
\midrule
% \rowcolor{tabgreenB}
VP & $\alpha_t = e^{-t}$ & $x_t = \sqrt{\alpha_t}\,x_0 + \sqrt{1-\alpha_t}\,\boldsymbol{\epsilon}$ & $d x_t = -\tfrac{1}{2}x_t\,dt + dW_t$ \\
% \rowcolor{tabgreenA}
VE-Karras & $\sigma$ & $z_\sigma = z_0 + \sigma\,\boldsymbol{\epsilon}$ & $d z_{\sigma} = \sqrt{2\sigma}\,dW_{\sigma}$ \\
% \rowcolor{tabgreenB}
Rectified flow & $s$ & $r_s = (1-s)\,r_0 + s\,\boldsymbol{\epsilon}$ & $d r_{s} = -\frac{r_s}{1-s}\,ds + \sqrt{\frac{2s}{1-s}}\,dW_{s}$ \\
\bottomrule
\end{tabular}%
}
\endgroup
\end{table}

\begin{figure}[t]
\centering
\includegraphics[width=0.95\linewidth]{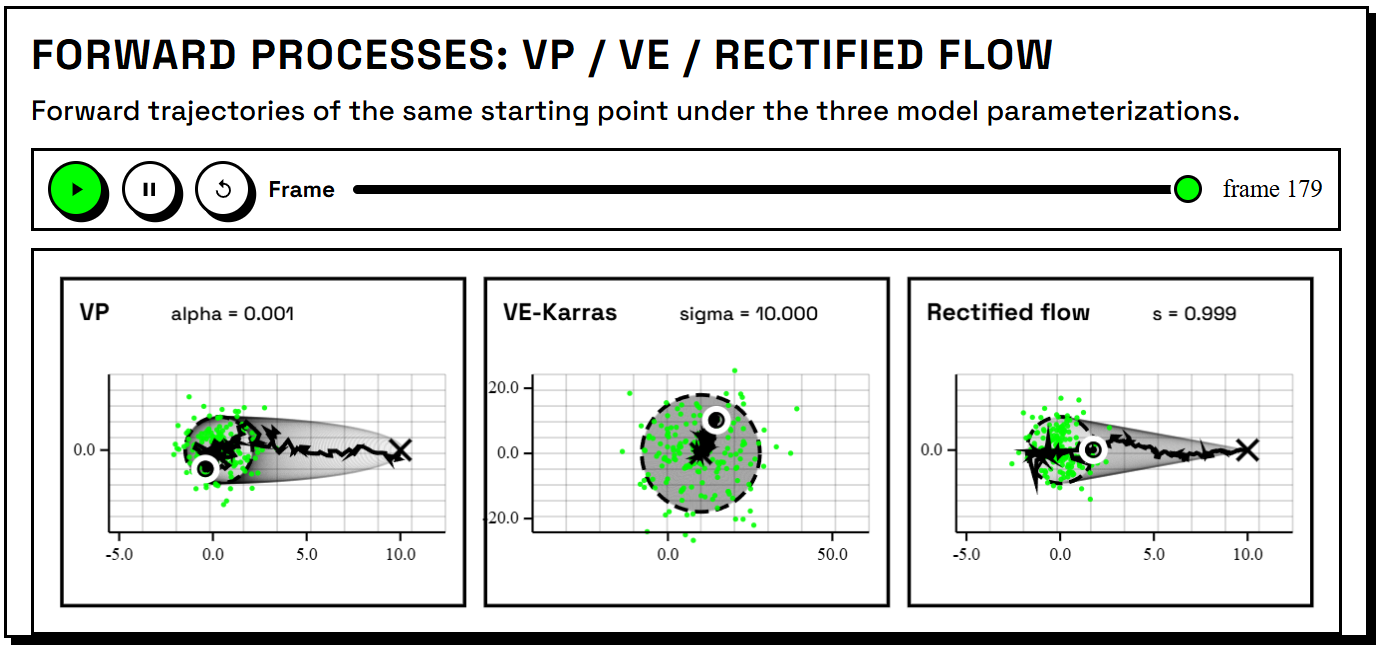}
\caption{Overview of forward processes across VP, VE-Karras, and rectified-flow parameterizations (exported from the original interactive visualization \url{https://iclr-blogposts.github.io/2026/blog/2026/rethinking-diffusion-langevin/}).}
\label{fig:forward-processes-overview}
\end{figure}

Each forward process has a characteristic way of mixing data and noise: the VP model uses the Ornstein--Uhlenbeck process, so samples drift toward the origin while their uncertainty grows; the VE-Karras model adds noise directly to the data without a restoring drift, so the mean stays fixed while the sample cloud expands outward; and the Rectified flow model is a stochastic forward process as well, not a deterministic straight-line interpolation. This behavior is illustrated in \cref{fig:forward-processes-overview}.

\begin{figure}[H]
\centering
\includegraphics[width=0.6\linewidth]{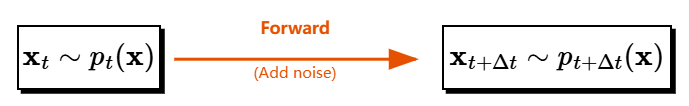}
\caption{A forward diffusion step with step size $\Delta t$ adds Gaussian noise to data, pushing samples closer to a Gaussian distribution.}
\label{fig:forward-step}
\end{figure}

Despite their differences, all above SDEs are fundamentally equivalent; they differ only by how time and state are reparameterized. For clarity, \cref{tab:forward-conversion} gives a direct conversion between any two parameterizations \citep{zheng2025lanpaint}. Using this table, one can directly translate between any two parameterizations whenever needed. No matter which notation we choose, a forward diffusion step with a step size of $\Delta t$ acts as adding more noise to data, which is displayed in \cref{fig:forward-step}.

\begin{table}[t]
\centering
\caption{Conversion between forward-process variable parameterizations.}
\label{tab:forward-conversion}
\begingroup
\renewcommand{\arraystretch}{1.5}
\setlength{\tabcolsep}{8pt}
\large

\definecolor{cellyellowdark}{RGB}{255,214,170}
\definecolor{cellyellowdarkdark}{RGB}{244,176,110}
\resizebox{\textwidth}{!}{%
\begin{tabular}{l!{\color{white}\vrule width 1.2pt}l!{\color{white}\vrule width 1.2pt}l!{\color{white}\vrule width 1.2pt}l}
\toprule
% \rowcolor{headerblue}
\textbf{Given parameterization} & \textbf{Equivalent VP } & \textbf{Equivalent VE-Karras } & \textbf{Equivalent Rectified-flow } \\
\midrule
\cellcolor{rowgray} \textbf{VP} $(x_t,\alpha_t)$ & \multicolumn{1}{c}{\cellcolor{white}\shortstack[c]{\raisebox{0.55em}{\textbf{/}}}} & \cellcolor{cellyellow}\shortstack[c]{$z_\sigma = \dfrac{x_t}{\sqrt{\alpha_t}}$\\[0.25em]$\sigma = \sqrt{\dfrac{1-\alpha_t}{\alpha_t}}$} & \cellcolor{cellyellowdark}\shortstack[c]{$r_s = \dfrac{x_t}{\sqrt{\alpha_t}+\sqrt{1-\alpha_t}}$\\[0.25em]$s = \dfrac{\sqrt{1-\alpha_t}}{\sqrt{\alpha_t}+\sqrt{1-\alpha_t}}$} \\
\addlinespace[0.2em]
\cellcolor{rowgray}\textbf{VE-Karras} $(z_\sigma,\sigma)$ & \multicolumn{1}{c}{\cellcolor{cellyellow}\shortstack[c]{$x_t = \dfrac{z_\sigma}{\sqrt{1+\sigma^2}}$\\[0.25em]$\alpha_t = \dfrac{1}{1+\sigma^2}$}} & \multicolumn{1}{c}{\cellcolor{white}\shortstack[c]{\raisebox{0.55em}{\textbf{/}}}} & \cellcolor{cellyellowdarkdark}\shortstack[c]{$r_s = \dfrac{z_\sigma}{1+\sigma}$\\[0.25em]$s = \dfrac{\sigma}{1+\sigma}$} \\
\addlinespace[0.2em]
\cellcolor{rowgray}\textbf{Rectified flow} $(r_s,s)$ & \cellcolor{cellyellowdark}\shortstack[c]{$x_t = \dfrac{r_s}{\sqrt{(1-s)^2+s^2}}$\\[0.25em]$\alpha_t = \dfrac{(1-s)^2}{(1-s)^2+s^2}$} & \cellcolor{cellyellowdarkdark}\shortstack[c]{$z_\sigma = \dfrac{r_s}{1-s}$\\[0.25em]$\sigma = \dfrac{s}{1-s}$} & \multicolumn{1}{c}{\cellcolor{white}\shortstack[c]{\raisebox{0.55em}{\textbf{/}}}} \\
\bottomrule
\end{tabular}%
}
\endgroup
\end{table}

\subsection{The Reverse Diffusion Process for Denoising}

The reverse diffusion process is the conjugate of the forward process. While the forward process evolves $p_t(\mathbf{x})$ toward Gaussian noise, the reverse process reverses this evolution, restoring Gaussian noise to $p_t$.

\begin{figure}[t]
\centering
\includegraphics[width=0.6\linewidth]{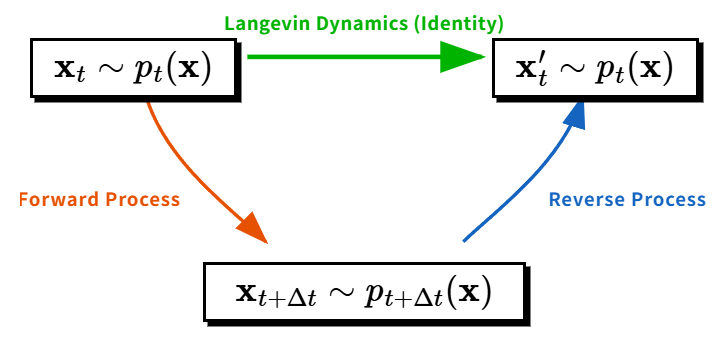}
\caption{The forward and reverse diffusion processes compose to reproduce Langevin dynamics.}
\label{fig:forward-reverse-split}
\end{figure}

The concept behind the reverse process is intuitive: since Langevin dynamics acts as an identity operation on a distribution---preserving it unchanged---any forward process composed with its corresponding reverse process should similarly yield a Langevin dynamics. Specifically, at any time $t$, combining the forward and reverse processes should reproduce the Langevin dynamics for the distribution $p_t(\mathbf{x})$, as illustrated in \cref{fig:forward-reverse-split}.

\begin{table}[t]
\begingroup
\renewcommand{\arraystretch}{1.5}
\setlength{\tabcolsep}{8pt}
\centering
\caption{Langevin split of different model types.}
\label{tab:langevin-split}
\resizebox{\textwidth}{!}{%
\begin{tabular}{llll}
\toprule
% \rowcolor{tabgreenA}
\textbf{Model Type} & \textbf{Langevin dynamics} & \textbf{Reverse Split} & \textbf{Forward Split} \\
\midrule
% \rowcolor{tabgreenB}
VP-SDE & \cellcolor{cellgreen}$dx = \mathbf{s}_x\,d\tau + \sqrt{2}\,dW_\tau$ & \cellcolor{cellblue}$dx = \left[\frac{1}{2}x + \mathbf{s}_x\right]d\tau + dW_{\tau}$ & \cellcolor{cellyellow}$dx = -\tfrac{1}{2}x\,d\tau + dW_\tau$ \\
% \rowcolor{tabgreenA}
VP-ODE & \cellcolor{cellgreen}$dx = \frac{1}{2}\mathbf{s}_x\,d\tau + dW_\tau$ & \cellcolor{cellblue}$dx = \frac{1}{2}\left(x + \mathbf{s}_x\right)d\tau$ & \cellcolor{cellyellow}$dx = -\tfrac{1}{2}x\,d\tau + dW_\tau$ \\
% \rowcolor{tabgreenB}
VE-Karras & \cellcolor{cellgreen}$dz = \tau\,\mathbf{s}_z\,d\tau + \sqrt{2\tau}\,dW_\tau$ & \cellcolor{cellblue}$dz = \tau\,\mathbf{s}_z\,d\tau$ & \cellcolor{cellyellow}$dz = \sqrt{2\tau}\,dW_{\tau}$ \\
% \rowcolor{tabgreenA}
Rectified flow & \cellcolor{cellgreen}$dr = \frac{\tau}{1+\tau}\mathbf{s}_r\,d\tau + \sqrt{\frac{2\tau}{1+\tau}}\,dW_\tau$ & \cellcolor{cellblue}$dr = \frac{\tau\,\mathbf{s}_r + r}{1-\tau}d\tau$ & \cellcolor{cellyellow}$dr = -\frac{r}{1-\tau}\,d\tau + \sqrt{\frac{2\tau}{1-\tau}}\,dW_{\tau}$ \\
\bottomrule
\end{tabular}%
}
\endgroup
\end{table}

To formalize this, consider the VP case with the following Langevin dynamics for $p_t(\mathbf{x})$ with a time variable $\tau$, distinguished from the forward diffusion time $t$. This dynamics can be decomposed into forward and reverse components as follows:
\begin{equation}
\begin{split}
d\mathbf{x}_\tau
&= \mathbf{s}(\mathbf{x}_\tau,t)\,d\tau + \sqrt{2}\,d\mathbf{W}_\tau \\
&= \underbrace{-\frac{1}{2}\mathbf{x}_\tau\,d\tau + d\mathbf{W}_\tau^{(1)}}_{\text{Forward}}
+ \underbrace{\left(\frac{1}{2}\mathbf{x}_\tau + \mathbf{s}(\mathbf{x}_\tau,t)\right)d\tau + d\mathbf{W}_\tau^{(2)}}_{\text{Reverse}}.
\end{split}
\label{eq:vp-split}
\end{equation}
where $\mathbf{s}(\mathbf{x}, t) = \nabla_{\mathbf{x}} \log p_t(\mathbf{x})$ is the score function of $p_t(\mathbf{x})$. Here, we split the noise term $\sqrt{2}\, d\mathbf{W}_\tau$ into two independent Gaussian increments, $d\mathbf{W}_\tau^{(1)}$ and $d\mathbf{W}_\tau^{(2)}$, such that their sum equals the original noise: $\sqrt{2}\, d\mathbf{W}_\tau = d\mathbf{W}_\tau^{(1)} + d\mathbf{W}_\tau^{(2)}$. This split is possible because Gaussian random variables satisfy the property that their sum is Gaussian, and independent Gaussians add in variance; specifically, if $d\mathbf{W}_\tau^{(1)}$ and $d\mathbf{W}_\tau^{(2)}$ are independent standard Brownian increments (each with variance $d\tau$), their sum has variance $2\,d\tau$, matching the original $\sqrt{2}\,d\mathbf{W}_\tau$.

\noindent This decomposition now lets us directly answer the first question posed in the abstract:
\begin{questionbox}
\textbf{How does the reverse process invert the forward process to generate data from pure noise?}
\end{questionbox}

The ``Forward'' part in this decomposition corresponds to the forward diffusion process, effectively increasing the forward diffusion time $t$ by $d\tau$, bringing the distribution to $p_{t + d\tau}(\mathbf{x})$. Since the forward and reverse components combine to form an ``identity'' Langevin dynamics, the ``Reverse'' part must reverse the forward process, decreasing the forward diffusion time $t$ by $d\tau$ and restoring the distribution back to $p_t(\mathbf{x})$.

We can therefore read off the reverse process as
\begin{equation}
d\mathbf{x}_{t'} = \left(\frac{1}{2}\mathbf{x}_{t'} + \mathbf{s}(\mathbf{x}_{t'},t)\right)dt' + d\mathbf{W}_{t'}.
\end{equation}
This reverse diffusion process is itself a standalone SDE that advances reverse time $t'$. If $\mathbf{x}_{t'}\sim q_{t'}(\mathbf{x})$, then a step with increment $dt'=\Delta t'$ moves it to $\mathbf{x}_{t'+\Delta t'}\sim q_{t'+\Delta t'}(\mathbf{x})$.

Having analyzed the VP case in detail, we can now apply the same decomposition approach to other diffusion schemes, which involve different choices of Langevin dynamics. This brings us to the second question raised in the abstract:

\begin{questionbox}
\textbf{How can ODE-based and SDE-based diffusion models be unified under a single framework?}
\end{questionbox}

\cref{tab:langevin-split} provides a direct answer: these models are unified by decomposing different Langevin dynamics. We have decomposed the VP model into both SDE and ODE versions, as well as other parameterizations, relating their Langevin dynamics to the corresponding forward and reverse processes.

A key observation from this table is that the Langevin split is not unique. For the same VP model, we present two distinct splittings, the SDE and ODE versions, which are decompositions of different Langevin dynamics differing in their time scaling functions $g(\tau)$. The ODE version corresponds to a splitting where the reverse process contains no stochastic term $dW$.

Besides the decomposition of Langevin dynamics, we still have one problem: note that the $\mathbf{s}(\mathbf{x}_{t'}, t)$ term in the reverse process still depends on the forward time $t$, not the reverse time $t'$; we need the relationship between the forward time $t$ and the reverse time $t'$ to close the equation. Note that a single reverse-time step $dt'$ can be understood in two complementary ways:
\begin{enumerate}[leftmargin=1.5em]
  \item As an undoing of the forward diffusion: one step of the reverse diffusion process with $dt' = \Delta t$ removes a small amount of noise and therefore reduces the forward diffusion time by $\Delta t$.
  \item As forward evolution in its own clock: the reverse diffusion process is itself a well-defined SDE/ODE in the variable $t'$, so one step with $dt' = \Delta t$ simply advances the reverse diffusion time from $t'$ to $t' + \Delta t$.
\end{enumerate}
Together, these two viewpoints determine how the forward and reverse clocks are related. Since a positive reverse-time step $dt' > 0$ both decreases the forward time $t$ and increases the reverse time $t'$, their infinitesimal increments must satisfy
\begin{equation}
dt = -dt'.
\end{equation}
which means that $t'$ runs in the opposite direction to $t$. To make $t'$ lie in the same range $[0, T]$ as the forward diffusion time, we can define
\begin{equation}
t = T - t'.
\end{equation}
so that $t = 0$ corresponds to $t' = T$ and $t = T$ corresponds to $t' = 0$. In this notation, the reverse diffusion process of VP is
\begin{equation}
d\mathbf{x}_{t'} = \left(\frac{1}{2}\mathbf{x}_{t'} + \mathbf{s}(\mathbf{x}_{t'},T-t')\right)dt' + d\mathbf{W}_{t'}.
\label{eq:reverse-process-vp}
\end{equation}
in which $t' \in [0,T]$ is the reverse time, $\mathbf{s}(\mathbf{x}, t) = \nabla_{\mathbf{x}} \log p_t(\mathbf{x})$ is the score function of the density of $\mathbf{x}_{t}$ in the forward process.

The same reasoning applies not only to SDE reverse processes but also to ODE reverse processes. The full summary is listed in \cref{tab:reverse-processes}.

\begin{table}[t]
\begingroup
\renewcommand{\arraystretch}{1.5}
\setlength{\tabcolsep}{8pt}
\centering
\caption{Reverse diffusion processes across model types.}
\label{tab:reverse-processes}
\resizebox{\textwidth}{!}{%
\begin{tabular}{lllll}
\toprule
% \rowcolor{tabgreenA}
\textbf{Model Type} & \textbf{Reverse Process} & \textbf{Relation to Score} & \textbf{Reverse Time} & \textbf{Reverse time domain} \\
\midrule
% \rowcolor{tabgreenB}
VP-SDE & $d\mathbf{x}_{t'} = \left[\frac{1}{2}\mathbf{x}_{t'} + \mathbf{s}(\mathbf{x}_{t'},T-t')\right]dt' + d\mathbf{W}_{t'}$ & $\mathbf{s}(\mathbf{x},t) = \mathbf{s}_x(\mathbf{x},t)$ & $t' = T - t$ & $t' \in [0,T]$ \\
% \rowcolor{tabgreenA}
VP-ODE & $d\mathbf{x}_{t'} = \frac{1}{2}\left[\mathbf{x}_{t'} + \mathbf{s}(\mathbf{x}_{t'},T-t')\right]dt'$ & $\mathbf{s}(\mathbf{x},t) = \mathbf{s}_x(\mathbf{x},t)$ & $t' = T - t$ & $t' \in [0,T]$ \\
% \rowcolor{tabgreenB}
VE-Karras & $d\mathbf{z}_{\sigma'} = -\boldsymbol{\epsilon}(\mathbf{z}_{\sigma'},\Sigma-\sigma')\,d\sigma'$ & $\boldsymbol{\epsilon}(\mathbf{z},\sigma) = -\sigma \mathbf{s}_z(\mathbf{z},\sigma)$ & $\sigma' = \Sigma-\sigma$ & $\sigma' \in [0,\Sigma]$ \\
% \rowcolor{tabgreenA}
Rectified flow & $d\mathbf{r}_{s'} = -\mathbf{v}(\mathbf{r}_{s'},1-s')\,ds'$ & $\mathbf{v}(\mathbf{r},s) = -\frac{s\,\mathbf{s}_r(\mathbf{r},s)+\mathbf{r}}{1-s}$ & $s' = 1-s$ & $s' \in [0,1]$ \\
\bottomrule
\end{tabular}%
}
\endgroup
\end{table}

In this table, $\boldsymbol{\epsilon}$ and $\mathbf{v}$ are just different ways of writing expressions based on the basic score functions. The score functions themselves are
\begin{equation}
\mathbf{s}_x(\mathbf{x},t) = \nabla_{\mathbf{x}_t}\log p(\mathbf{x}_t),\qquad
\mathbf{s}_z(\mathbf{z},\sigma) = \nabla_{\mathbf{z}_\sigma}\log p(\mathbf{z}_\sigma),\qquad
\mathbf{s}_r(\mathbf{r},s) = \nabla_{\mathbf{r}_s}\log p(\mathbf{r}_s).
\end{equation}

\begin{figure}[t]
\centering
\includegraphics[width=0.95\linewidth]{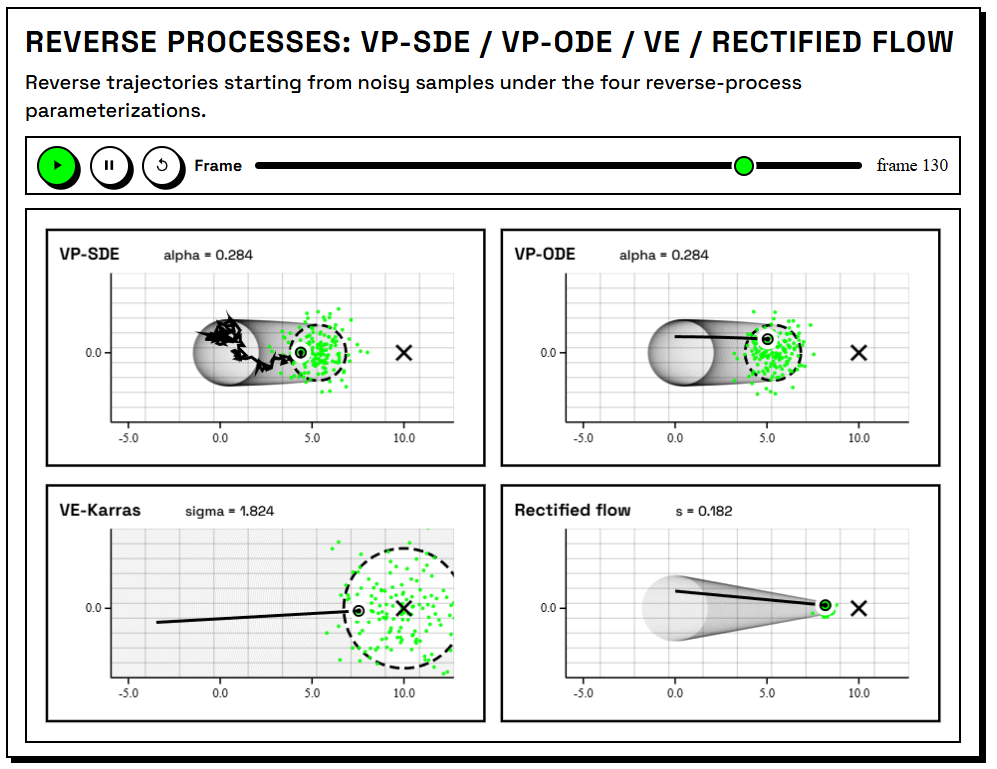}
\caption{Reverse trajectories under different parameterizations (exported from the original interactive visualization \url{https://iclr-blogposts.github.io/2026/blog/2026/rethinking-diffusion-langevin/}).}
\label{fig:reverse-processes-overview}
\end{figure}

These reverse equations become more intuitive when we visualize how samples move under each parameterization, as shown in \cref{fig:reverse-processes-overview}:

\begin{table}[t]
\begingroup
\renewcommand{\arraystretch}{1.75}
\setlength{\tabcolsep}{8pt}
\large
\definecolor{cellblueA}{RGB}{232,242,255}
\definecolor{cellblueB}{RGB}{206,226,252}
\definecolor{cellblueC}{RGB}{176,208,246}
\centering
\caption{Conversion between model prediction.}
\label{tab:native-prediction-conversion}
\resizebox{\textwidth}{!}{%
\begin{tabular}{l!{\color{white}\vrule width 1.2pt}l!{\color{white}\vrule width 1.2pt}l!{\color{white}\vrule width 1.2pt}l}
\toprule
\textbf{Given prediction} & \textbf{Equivalent VP score $\mathbf{s}_x$} & \textbf{Equivalent VE noise $\boldsymbol{\epsilon}$} & \textbf{Equivalent RF velocity $\mathbf{v}$} \\
\midrule
\cellcolor{rowgray}\textbf{VP score} $\mathbf{s}_x(x_t,\alpha_t)$ & \multicolumn{1}{c}{\cellcolor{white}\shortstack[c]{\raisebox{0.55em}{\textbf{/}}}} & \cellcolor{cellblueA}$\boldsymbol{\epsilon}(z_\sigma,\sigma) = -\sqrt{1-\alpha_t}\,\mathbf{s}_x(x_t,\alpha_t)$ & \cellcolor{cellblueB}$\mathbf{v}(r_s,s) = -\frac{x_t}{\sqrt{\alpha_t}} - \frac{1-\alpha_t+\sqrt{\alpha_t(1-\alpha_t)}}{\sqrt{\alpha_t}}\,\mathbf{s}_x(x_t,\alpha_t)$ \\
\addlinespace[0.2em]
\cellcolor{rowgray}\textbf{VE noise} $\boldsymbol{\epsilon}(z_\sigma,\sigma)$ & \cellcolor{cellblueA}$\mathbf{s}_x(x_t,\alpha_t) = -\frac{\sqrt{1+\sigma^2}}{\sigma}\,\boldsymbol{\epsilon}(z_\sigma,\sigma)$ & \multicolumn{1}{c}{\cellcolor{white}\shortstack[c]{\raisebox{0.55em}{\textbf{/}}}} & \cellcolor{cellblueC}$\mathbf{v}(r_s,s) = (1+\sigma)\boldsymbol{\epsilon}(z_\sigma,\sigma)-z_\sigma$ \\
\addlinespace[0.2em]
\cellcolor{rowgray}\textbf{RF velocity} $\mathbf{v}(r_s,s)$ & \cellcolor{cellblueB}$\mathbf{s}_x(x_t,\alpha_t) = -\frac{\sqrt{(1-s)^2+s^2}}{s}\left(r_s+(1-s)\mathbf{v}(r_s,s)\right)$ & \cellcolor{cellblueC}$\boldsymbol{\epsilon}(z_\sigma,\sigma) = r_s+(1-s)\mathbf{v}(r_s,s)$ & \multicolumn{1}{c}{\cellcolor{white}\shortstack[c]{\raisebox{0.55em}{\textbf{/}}}} \\
\bottomrule
\end{tabular}%
}
\endgroup
\end{table}

In this single-data-point example, the reverse trajectories reveal a clear geometric difference between the parameterizations. The VP-SDE and VP-ODE flows bend along a curved path as they return to the target point, whereas the VE-Karras and Rectified flow trajectories move approximately along a straight line toward that point. It is important to emphasize that this straight-line behavior is a special feature of the one-point setting shown in the example, not the general case. For a general data distribution, the learned reverse vector fields vary across space, so all of these reverse trajectories are typically curved. Nevertheless, one could still expect the VE-Karras and Rectified flow trajectories to have smaller curvature than the VP trajectories.

\subsection{Converting Between Different Model Types}

Despite their different geometric behaviors, all model types we discussed above are inherently equivalent parameterizations. Although VP uses the score $\mathbf{s}_x$, VE-Karras uses the noise prediction $\boldsymbol{\epsilon}$, and Rectified flow uses the velocity field $\mathbf{v}$ as their native outputs, these model types are mathematically equivalent parameterizations. Combined with the previous conversion table for the forward-process variables, we can therefore convert these fields into one another exactly \citep{zheng2025lanpaint}.

\Cref{tab:native-prediction-conversion} summarizes these conversions. From this table, we can see directly that the velocity learned in flow matching is equivalent to the noise prediction and the score under a change of parameterization. Its main advantage is therefore not that it produces truly straight-line trajectories, but that it is often expected to produce trajectories with smaller curvature.

\section{Forward--Reverse Duality}

We have established that a single reverse step undoes a forward step: advancing the reverse time $t'$ by an amount corresponds to decreasing the forward time $t$ by the same amount. Now, let us examine what happens when we combine multiple forward and reverse steps to reveal the deeper duality between them. In fact, the forward process transforms a data distribution into noise, while the reverse process, starting from noise, generates samples from the same data distribution.

\begin{figure}[t]
\centering
\includegraphics[width=0.7\linewidth]{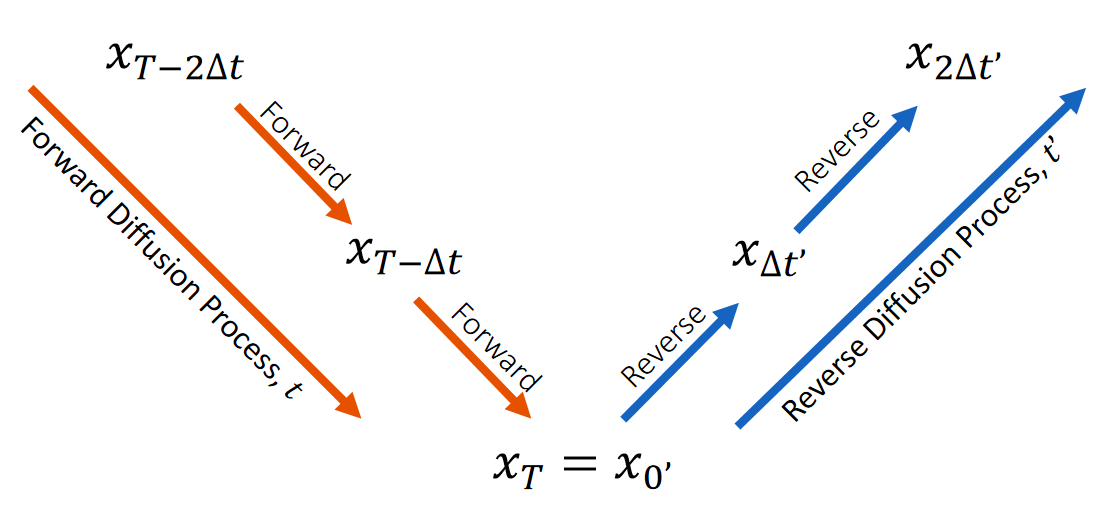}
\caption{Part of a forward--reverse diffusion cycle: the last two steps of the forward process (green arrows, increasing $t$) followed by the first two steps of the reverse process (blue arrows, increasing $t'$ while decreasing $t$).}
\label{fig:forward-reverse-cycle-part}
\end{figure}

Consider the following sequence: begin with a data sample $\mathbf{x}_0$, propagate it through the forward process to obtain $\mathbf{x}_T$, then use $\mathbf{x}_T$ as the starting point $\mathbf{x}_{0'}$ for the reverse process and evolve it to $\mathbf{x}_{T'}$. Part of this forward--reverse cycle is illustrated in \cref{fig:forward-reverse-cycle-part}.

\begin{figure}[t]
\centering
\includegraphics[width=0.8\linewidth]{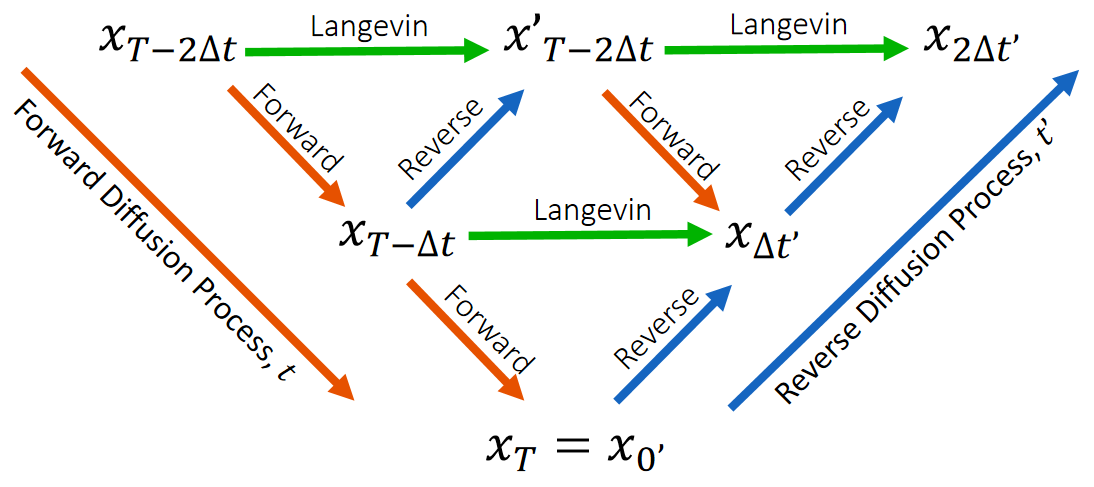}
\caption{Each horizontal row shows a Langevin dynamics step that maps a forward sample $\mathbf{x}_t$ to a new reverse sample $\mathbf{x}_{(T-t)'}$ from the same probability density.}
\label{fig:forward-reverse-duality-rows}
\end{figure}

The green arrows represent consecutive forward process steps that advance the forward diffusion time $t$, while the blue arrows indicate consecutive reverse process steps that advance the reverse diffusion time $t'$. We examine the relationship between $\mathbf{x}_{t}$ in the forward diffusion process and $\mathbf{x}_{t'=T-t}$ in the reverse diffusion process. The composition of a forward and a reverse step constitutes a Langevin dynamics step. This allows us to connect $\mathbf{x}$ in the forward process with those in the reverse process through Langevin dynamics steps, as illustrated in \cref{fig:forward-reverse-duality-rows}.

Each horizontal row in this picture corresponds to consecutive steps of Langevin dynamics, which alter the samples while maintaining the same probability density. This illustrates the \textbf{duality} between the forward and reverse diffusion processes: while $\mathbf{x}_t$ (forward) and $\mathbf{x}_{(T-t)'}$ (reverse) are distinct samples, they obey the same probability distribution.

To formalize the duality, let $p_t(\mathbf{x})$ denote the density of the forward process at time $t$, and let $q_{t'}(\mathbf{x})$ denote the density of the reverse process at reverse time $t'$. If we initialize
\begin{equation}
q_0(\mathbf{x}) = p_T(\mathbf{x}),
\end{equation}
then their evolution are related by
\begin{equation}
q_{t'}(\mathbf{x}) = p_{T-t'}(\mathbf{x}).
\end{equation}

In diffusion models, the terminal time $T$ is chosen sufficiently large that the forward-process distribution $p_T(\mathbf{x})$ converges to a simple Gaussian distribution. This ensures that the reverse process can start from the same Gaussian distribution $q_0(\mathbf{x})$ at $t'=0$. By then evolving the reverse process through time $t'$ from $0$ to $T$, we obtain samples that follow the original data distribution:
\begin{equation}
q_T(\mathbf{x}) = p_0(\mathbf{x}) \qquad \text{(data distribution)}.
\end{equation}

This exact recovery of the data distribution $p_0$ through a forward--reverse duality brings us to the third question from the abstract.

\begin{questionbox}
\textbf{Why are diffusion models theoretically superior to ordinary VAEs?}
\end{questionbox}

The above duality means that if we run the reverse process from time $t'=0$ to $t'=T$, the final samples follow exactly the same distribution as the original training data $p_0$. In other words, the forward and reverse processes form an exact prior--posterior pair: the forward process maps data to noise, and the reverse process maps noise back to data. In practice, training introduces approximation error, but the theoretical target is exact equality. Ordinary VAEs, by contrast, only require the decoder to approximate the encoder posterior, with no guarantee of exactness even at the ELBO optimum.

Now we have demonstrated that \textbf{reverse diffusion}---the dual of the forward process---can generate image data from noise. However, this requires access to the score function at every time step $t$. In practice, we approximate this function using a neural network. In the next section, we will explain how to train such score networks.

\section{Unifying Training of Diffusion Models as Maximal likelihood}

In this section, we derive the training objective directly from the maximum-likelihood framework. By doing so, we reveal the fundamental connection between diffusion model loss and exact maximum likelihood, and show that score matching, denoising, and flow matching are equivalent manifestations of this same objective rather than fundamentally different levels of simplicity.

Training the diffusion model involves addressing two fundamental questions: (1) What mathematical quantity should we model, and (2) What objective function should guide the training? Here, we start by analyzing the Kullback--Leibler (KL) divergence.

Suppose we have two distributions $p(\mathbf{x},t)$ and $q(\mathbf{x},t)$ that both evolve under the same forward diffusion process. Think of $p$ as the true data distribution pushed forward by the diffusion dynamics, and $q$ as the model distribution. At any fixed time $t$, their KL divergence is
\begin{equation}
\KL\!\left(p_t\|q_t\right) = \int p(\mathbf{x},t)\log\frac{p(\mathbf{x},t)}{q(\mathbf{x},t)}\,d\mathbf{x}.
\end{equation}

Maximum likelihood training aims to minimize the KL divergence $\KL(p_0\|q_0)$ at time $t=0$, where $p_0$ is the true data distribution and $q_0$ is the model distribution. However, in diffusion models, we introduce a forward process that evolves distributions over time $t$, and we learn a reverse process that maps from noisy states at different times back to clean data. This temporal structure suggests that rather than focusing solely on the KL divergence at $t=0$, we should consider how this divergence evolves throughout the entire diffusion process. The key insight is to distribute the KL minimization objective across all diffusion times by examining the time derivative of $\KL(p_t\|q_t)$ along the forward dynamics.

Formally, we can rewrite the time-zero KL as an integral over its time derivative:
\begin{equation}
\begin{aligned}
\KL\!\left(p_0\|q_0\right)
&= \KL\!\left(p_0\|q_0\right) - \KL\!\left(p_\infty\|q_\infty\right) \\
&= -\int_0^\infty \frac{d}{dt}\KL\!\left(p_t\|q_t\right)\,dt,
\end{aligned}
\end{equation}
where the second equality uses $\KL(p_\infty\|q_\infty)=0$ at infinitely large time, since both $p$ and $q$ converge to the same Gaussian noise distribution.

This naturally identifies the \emph{instantaneous contribution} to the likelihood objective as
\begin{equation}
L_t := -\frac{d}{dt}\KL\!\left(p_t\|q_t\right).
\end{equation}
Thus minimizing $\KL(p_0\|q_0)$ is equivalent to minimizing these contributions on average over diffusion time.

We now show that as long as the forward diffusion process takes the form
\begin{equation}
d\mathbf{x} = f(\mathbf{x},t)\,dt + g(t)\,d\mathbf{W},
\end{equation}
the instantaneous contribution is
\begin{equation}
\begin{aligned}
L_t
  &= \frac{1}{2} g(t)^2
     \int p(\mathbf{x}, t)\,
          \big\|\nabla \log p(\mathbf{x}, t)
                - \nabla \log q(\mathbf{x}, t)\big\|^2
        d\mathbf{x} \\
  &= \frac{1}{2} g(t)^2 \E_{\mathbf{x}\sim p(\mathbf{x},t)}
            \big\|\nabla \log p(\mathbf{x}, t)
                  - \nabla \log q(\mathbf{x}, t)\big\|^2.
\end{aligned}
\label{eq:score-objective}
\end{equation}

\Cref{eq:score-objective} shows that the score functions $\nabla \log p(\mathbf{x}, t)$ and $\nabla \log q(\mathbf{x}, t)$ for the true data distribution and the model distribution appear naturally inside the objective. Hence, the score function naturally arises as the quantity we should model. Full derivations of the Fokker--Planck equation and KL decay are provided in \cref{sec:appendix-fp,sec:appendix-kl}.

In practice, we approximate the model score $\nabla\log q(\mathbf{x},t)$ using a neural network. For standard score-based models, we model $\mathbf{s}_\theta(\mathbf{x},t)$ directly. For VE-Karras and rectified-flow parameterizations, we instead model related quantities such as noise prediction $\boldsymbol{\epsilon}$ or velocity $\mathbf{v}$, which can be converted back to a score.

The only thing remains to handle is the score of the true data distribution $\nabla\log p(\mathbf{x},t)$, which should be approximated by an empirical value from samples since we do not know its value. In fact,
\begin{equation}
\text{argmin}_{\mathbf{s}_\theta}
\E_{\mathbf{x}_0\sim p_0}
\E_{\mathbf{x}_t\sim p_t(\cdot\mid \mathbf{x}_0)}
\left\|\nabla \log p(\mathbf{x}_t | \mathbf{x}_0)-\mathbf{s}_\theta\right\|^2
= \text{argmin}_{\mathbf{s}_\theta}
\E_{\mathbf{x}\sim p(\mathbf{x},t)}
\left\|\nabla \log p(\mathbf{x},t)-\mathbf{s}_\theta\right\|^2.
\end{equation}
The left-hand side is the denoising score matching loss, while the right-hand side is the score matching loss. Their equivalence is shown in \cref{sec:appendix-dsm-sm}.

This tells us that training the diffusion model, we only need to figure out the $\nabla \log p(\mathbf{x}_t \mid \mathbf{x}_0)$, then minimize the loss
\begin{equation}
L_t = \frac{1}{2}g(t)^2
\E_{\mathbf{x}_0 \sim p_0}\,
\E_{\mathbf{x}_t \sim p_t(\cdot \mid \mathbf{x}_0)}
\left\|
\nabla \log p(\mathbf{x}_t \mid \mathbf{x}_0) - \mathbf{s}_\theta
\right\|^2.
\end{equation}

Equipped with this instantaneous maximum-likelihood objective, we can now address the fourth and final question from the abstract.

\begin{questionbox}
\textbf{Why flow matching is not fundamentally simpler than denoising or score matching, but equivalent under maximum-likelihood?}
\end{questionbox}

\begin{table}[t]
\centering
\caption{Training targets and losses under different parameterizations.}
\label{tab:training-parameterizations}
\begingroup
\small
\setlength{\tabcolsep}{5pt}
\renewcommand{\arraystretch}{1.3}
\resizebox{\textwidth}{!}{%
\begin{tabular}{lccccc}
\toprule
\textbf{Model Type} & \textbf{Noise-state relation} & \textbf{Network output} & \textbf{$\mathbf{s}_\theta$ w.r.t. NN} & \textbf{$\nabla \log p(x_t \mid x_0)$} & \textbf{Loss $L_t$} \\
\midrule
VP & $x_t = \sqrt{\alpha_t}\,x_0 + \sqrt{1-\alpha_t}\,\boldsymbol{\epsilon}$ & $\mathbf{s}_{\theta}(x_t,t)$ & $\mathbf{s}_{\theta}(x_t,t)$ & $-\frac{\boldsymbol{\epsilon}}{\sqrt{1-\alpha_t}}$ & $\frac{1}{2}\E_{\mathbf{x}_0 \sim p_0}\E_{\mathbf{x}_t \sim p_t(\cdot \mid \mathbf{x}_0)}\left\|-\frac{\boldsymbol{\epsilon}}{\sqrt{1-\alpha_t}} - \mathbf{s}_{\theta}(x_t,t)\right\|^2$ \\
VE-Karras & $z_\sigma = z_0 + \sigma\,\boldsymbol{\epsilon}$ & $\boldsymbol{\epsilon}_{\theta}(z_\sigma,\sigma)$ & $-\frac{\boldsymbol{\epsilon}_{\theta}(z_\sigma,\sigma)}{\sigma}$ & $-\frac{\boldsymbol{\epsilon}}{\sigma}$ & $\frac{1}{\sigma}\E_{\mathbf{z}_0 \sim p_0}\E_{\mathbf{z}_\sigma \sim p_\sigma(\cdot \mid \mathbf{z}_0)} \left\| \boldsymbol{\epsilon}_{\theta}(z_\sigma,\sigma) - \boldsymbol{\epsilon} \right\|^2$ \\
Rectified flow & $r_s = (1-s)\,r_0 + s\,\boldsymbol{\epsilon}$ & $\mathbf{v}_{\theta}(r_s,s)$ & $\frac{-\mathbf{v}_{\theta}(r_s,s)(1-s)-r_s}{s}$ & $-\frac{\boldsymbol{\epsilon}}{s}$ & $\frac{1-s}{s}\E_{\mathbf{r}_0 \sim p_0}\E_{\mathbf{r}_s \sim p_s(\cdot \mid \mathbf{r}_0)}\left\| \boldsymbol{\epsilon}-r_0-\mathbf{v}_{\theta}(r_s,s) \right\|^2$ \\
\bottomrule
\end{tabular}%
}
\endgroup
\end{table}

With the maximum-likelihood objective derived above, we can compare different parameterizations in a common framework and see explicitly why flow matching is not a fundamentally simpler alternative, but an equivalent reformulation of denoising and score matching.

\cref{tab:training-parameterizations} shows the loss functions for different diffusion model types. For the VP model, the loss directly trains a score function. For the VE-Karras model, the loss trains a network $\boldsymbol{\epsilon}_\theta$ to predict the Gaussian noise added to the data; this is the familiar epsilon-prediction parameterization. Other choices such as $x_0$-prediction or $v$-prediction are algebraically equivalent reformulations of the same objective.

For the rectified-flow model, it looks like learning a constant velocity, but that is not the case. Note that with $r_s=(1-s)\,r_0+s\,\boldsymbol{\epsilon}$ we have $r_1=\boldsymbol{\epsilon}$, so the loss can be written as
\begin{equation}
\left\|r_1-r_0-\mathbf{v}_{\theta}(r_s,s)\right\|^2.
\end{equation}
If we interpret $r_0$ and $r_1$ as particle positions at times $s=0$ and $s=1$, then $r_1-r_0$ is the average velocity over $[0,1]$, which motivates viewing $\mathbf{v}_\theta$ as a velocity field and writing the reverse process as $dr = -\mathbf{v}(r,s)\, ds$. This has led to the intuition that rectified flows are trained on simple straight lines and are therefore conceptually simpler than diffusion models. However, $\mathbf{v}_\theta(r,s)$ still depends on time $s$, so the velocity changes over time and trajectories are not truly straight in state--time space. More importantly, \cref{tab:training-parameterizations} shows that this velocity field is algebraically tied to the same underlying score function that appears in denoising and score matching. Under the maximum-likelihood objective, flow matching is therefore best understood not as a fundamentally simpler class, but as an equivalent parameterization of the same diffusion objective.

A note on loss weighting is also important. In practice, the coefficient outside the $L_2$ norm, such as $\frac{1}{2}$, $\frac{1}{\sigma}$, or $\frac{1-s}{s}$, is often omitted or replaced with a custom weighting schedule to improve training performance. This is valid because modifying this coefficient only changes the relative importance of the loss across different time steps $t$---it does not affect the optimal solution at any individual time $t$. In other words, reweighting adjusts how much we prioritize learning at different noise levels, but the target (the true score or velocity) remains unchanged.

Combining all results from previous discussion, we summarize the forward, reverse, and loss for each diffusion type in \cref{tab:unified-summary}.

\begin{table}[t]
\begingroup
\renewcommand{\arraystretch}{1.5}
\setlength{\tabcolsep}{8pt}
\centering
\caption{Unified summary of forward process, reverse process, and objective.}
\label{tab:unified-summary}
\resizebox{\textwidth}{!}{%
\begin{tabular}{llll}
\toprule
\textbf{Model Type} & \textbf{Forward Process} & \textbf{Reverse Process} & \textbf{Loss (up to a weight factor)} \\
\midrule
VP-SDE & $x_t = \sqrt{\alpha_t}\,x_0 + \sqrt{1-\alpha_t}\,\boldsymbol{\epsilon}$ & $d x_{t'} = \left[\frac{1}{2}x_{t'}+\mathbf{s}(x_{t'},T-t')\right]dt' + dW_{t'}$ & $\E_{\mathbf{x}_0 \sim p_0}\E_{\mathbf{x}_t \sim p_t(\cdot \mid \mathbf{x}_0)}\left\|-\frac{\boldsymbol{\epsilon}}{\sqrt{1-\alpha_t}}-\mathbf{s}_{\theta}(x_t,t)\right\|^2$ \\
VP-ODE & $x_t = \sqrt{\alpha_t}\,x_0 + \sqrt{1-\alpha_t}\,\boldsymbol{\epsilon}$ & $d x_{t'} = \frac{1}{2}\left[x_{t'}+\mathbf{s}(x_{t'},T-t')\right]dt'$ & $\E_{\mathbf{x}_0 \sim p_0}\E_{\mathbf{x}_t \sim p_t(\cdot \mid \mathbf{x}_0)}\left\|-\frac{\boldsymbol{\epsilon}}{\sqrt{1-\alpha_t}}-\mathbf{s}_{\theta}(x_t,t)\right\|^2$ \\
VE-Karras & $z_\sigma = z_0 + \sigma\,\boldsymbol{\epsilon}$ & $d z_{\sigma'} = -\boldsymbol{\epsilon}(z_{\sigma'},\Sigma-\sigma')\,d \sigma'$ & $\E_{\mathbf{z}_0 \sim p_0}\E_{\mathbf{z}_\sigma \sim p_\sigma(\cdot \mid \mathbf{z}_0)}\left\| \boldsymbol{\epsilon}_{\theta}(z_\sigma,\sigma) - \boldsymbol{\epsilon} \right\|^2$ \\
Rectified flow & $r_s = (1-s)\,r_0 + s\,\boldsymbol{\epsilon}$ & $d r_{s'} = -\mathbf{v}(r_{s'},1-s')\,d s'$ & $\E_{\mathbf{r}_0 \sim p_0}\E_{\mathbf{r}_s \sim p_s(\cdot \mid \mathbf{r}_0)}\left\| \boldsymbol{\epsilon}-r_0-\mathbf{v}_{\theta}(r_s,s) \right\|^2$ \\
\bottomrule
\end{tabular}%
}
\endgroup
\end{table}

\section{Conclusion}

From the Langevin perspective, diffusion models become conceptually simple: the forward and reverse processes are just a carefully chosen split of Langevin dynamics, which itself is an ``identity map''. This viewpoint simultaneously explains how sampling inverts noising, unifies SDE and ODE formulations as different splittings of the same dynamics, and clarifies why diffusion models implement exact maximum likelihood in a way ordinary VAEs do not.

It also shows why flow matching is not fundamentally simpler than denoising or score matching, but instead an equivalent way of estimating the same underlying score field under the maximum-likelihood objective that governs Langevin dynamics. We hope this perspective helps demystify diffusion models to learners, so that new variants can be understood not as disconnected tricks, but as different parameterizations and discretizations of a single, coherent Langevin story.

\section*{Acknowledgements}
This work was supported in part by the General Research Fund 16302823, an Area of Excellence project (AoE/E-601/24-N), and a Theme-based Research Project (T32-615/24-R) from the Research Grants Council of the Hong Kong Special Administrative Region, China. We also acknowledge funding from the Hong Kong Innovation and Technology Commission (ITCPD/17-9).

\paragraph{Appendix.}
All optional derivations from the original blog are migrated to \cref{sec:appendix-stationary,sec:appendix-fp,sec:appendix-kl,sec:appendix-dsm-sm}.

\appendix
\section{Optional Derivations}

\subsection{Why $p(\mathbf{x})$ is stationary under Langevin dynamics}
\label{sec:appendix-stationary}

\begin{enumerate}[leftmargin=1.5em]
  \item Set $g(t) = 1$ by rescaling time as $t' = \int_0^t g(\tau)\, d\tau$. Under this change of variables, the dynamics become
  \[
  d\mathbf{x}_{t'} = \mathbf{s}(\mathbf{x}_{t'})\, dt' + \sqrt{2}\, d\mathbf{W}_{t'},
  \]
  which is equivalent to the case $g(t') = 1$. Thus, $g(t)$ only sets the time unit and does not affect the stationary distribution.

  \item Let us consider the dynamics in energy form as
  \[
  d\mathbf{x}_t = -\nabla E(\mathbf{x})\,dt + \sqrt{2}\,d\mathbf{W}_t.
  \]
  The random term $d\mathbf{W}_t$'s role is to perturb the system into complete, uniform chaos. The only position information is injected by the energy $E(\mathbf{x})$. Thus, the stationary distribution shall have the form $p(\mathbf{x}) = f(E(\mathbf{x}))$ for some function $f$.

  \item Consider $N$ independent copies $\mathbf{x}_1, \dots, \mathbf{x}_N$. Their joint density must be the product form $f(E(\mathbf{x}_1)) \cdots f(E(\mathbf{x}_N))$. From another point of view, when treating them as a single system, the total energy is additive:
  \[
  E(\mathbf{x}_1, \dots, \mathbf{x}_N) = \sum E(\mathbf{x}_i).
  \]
  Therefore, the joint stationary density of $N$ independent copies must also be the addition form $g(\sum E(\mathbf{x}_i))$ for some function $g$. The only function $f$ that turns product form into addition form is the exponential: $f(E) = e^{-\beta E}$. This yields
  \[
  p(\mathbf{x}) \propto e^{-\beta E(\mathbf{x})}.
  \]

  \item To find $\beta$, take $E(\mathbf{x}) = \frac{1}{2} \|\mathbf{x}\|^2$. This gives the well known Ornstein--Uhlenbeck process
  \[
  d\mathbf{x}_t = -\mathbf{x}\,dt + \sqrt{2}\,d\mathbf{W}_t
  \]
  with known stationary $\mathcal{N}(0, I)$, density $\propto e^{-\frac{1}{2} \|\mathbf{x}\|^2}$. Matching forms gives $\beta = 1$.
\end{enumerate}

Thus, the dynamics
\[
d\mathbf{x}_t = -\nabla E(\mathbf{x})\,dt + \sqrt{2}\,d\mathbf{W}_t
\]
has stationary distribution $\propto e^{-E(\mathbf{x})}$, and
\[
d\mathbf{x}_t = \nabla_{\mathbf{x}} \log p(\mathbf{x}) \, dt + \sqrt{2} \, d\mathbf{W}_t
\]
has stationary distribution $p(\mathbf{x})$.

\subsection{Derivation Step 1: from forward SDE to Fokker--Planck}
\label{sec:appendix-fp}

Given the SDE
\[
d\mathbf{x} = f(\mathbf{x}, t) \, dt + g(t) \, d\mathbf{W},
\]
we first ask: \textbf{how does the probabilistic density $p_t(\mathbf{x})$ evolve in time?} The answer is the \textbf{Fokker--Planck equation}, which describes the time evolution of the probability density $p(\mathbf{x}, t)$ induced by the SDE:
\[
\frac{\partial p}{\partial t}
:= -\nabla \cdot \left[f(\mathbf{x}, t)\, p\right]
  + \frac{1}{2}g(t)^2\, \nabla^2  p .
\]

This PDE shows how \textbf{drift} $f$ and \textbf{diffusion} $g$ jointly shape the distribution. Rigorous derivations can be found in standard references; here we only sketch an intuitive 1D argument for the drift part:

\textbf{Drift term $f$.} Start with a 1D motion with \textbf{constant velocity} $v$, so $dx = v\,dt$. After time $t$, a particle now at position $x$ must have come from $x - vt$ at time $0$, so
\[
p(x, t) = p(x - vt, 0).
\]

Differentiating this identity w.r.t. $t$ gives the continuity equation
\[
\frac{\partial p}{\partial t}
  + \frac{\partial}{\partial x}\big(v\, p(x, t)\big) = 0.
\]

For a general 1D deterministic dynamics $dx = f(x, t)\,dt$, the same reasoning yields
\[
\frac{\partial p}{\partial t}
  + \frac{\partial}{\partial x}\big(f(x, t)\, p(x, t)\big) = 0.
\]

We keep $f(x, t)$ inside the $\partial_x$ because this term represents the probability flux. This guarantees conservation: integrating the total derivative $\partial_x(f p)$ over all space gives zero (assuming $p$ vanishes at boundaries), preserving the total probability.

\textbf{Noise term $g\,dW$.} Consider now the pure diffusion SDE $dx = g\,dW$ with constant $g$ and initial condition $x(0) = 0$. At time $t$, the accumulated Brownian motion from $0$ to $t$ is Gaussian with variance $t$, so $x(t)$ is Gaussian with variance $g^2 t$ and density
\[
p(x, t) = \frac{1}{\sqrt{2\pi g^2 t}} \exp\!\left(-\frac{x^2}{2 g^2 t}\right).
\]

One can check directly that this density satisfies the diffusion equation
\[
\frac{\partial p}{\partial t} - \frac{1}{2} g^2 \frac{\partial^2 p}{\partial x^2} = 0.
\]

Combining drift and diffusion, we obtain that
\[
\frac{\partial p}{\partial t}
  = -\frac{\partial}{\partial x} \left[f(x, t)\, p\right]
    + \frac{1}{2} g(t)^2 \frac{\partial^2 p}{\partial x^2},
\]
which is the 1D specialization of the Fokker--Planck equation stated above.

\subsection{Derivation Step 2: KL decay and squared-score objective}
\label{sec:appendix-kl}

We now analyze \textbf{how the KL divergence between two solutions of the same Fokker--Planck equation evolves in time}.

Assume that both $p(\mathbf{x}, t)$ and $q(\mathbf{x}, t)$ satisfy the same Fokker--Planck equation with drift $f(\mathbf{x}, t)$ and diffusion strength $g(t)$:
\[
\frac{\partial p}{\partial t}
:= -\nabla \cdot \big(f p\big)
  + \frac{1}{2} g(t)^2 \nabla^2 p,
\qquad
\frac{\partial q}{\partial t}
:= -\nabla \cdot \big(f q\big)
  + \frac{1}{2} g(t)^2 \nabla^2 q.
\]

Define
\[
\mathrm{KL}\big(p_t \Vert q_t\big)
:= \int p(\mathbf{x}, t)\,\log\frac{p(\mathbf{x}, t)}{q(\mathbf{x}, t)}\,d\mathbf{x}.
\]

\textbf{Step 1: Differentiate the KL.} Differentiating under the integral sign and using $\int \partial_t p\,d\mathbf{x}=0$ (mass conservation), we obtain
\[
\frac{d}{dt}\mathrm{KL}\big(p_t \Vert q_t\big)
:= \int \left(\log\frac{p}{q}\right)\partial_t p\,d\mathbf{x}
  - \int \frac{p}{q}\,\partial_t q\,d\mathbf{x}.
\]

Introduce the Fokker--Planck operator
\[
\mathcal{L}u = -\nabla\cdot(fu) + \frac{1}{2}g(t)^2\nabla^2 u,
\]
so that $\partial_t p = \mathcal{L}p$ and $\partial_t q = \mathcal{L}q$. Let $r = p/q$. Then
\[
\frac{d}{dt}\mathrm{KL}\big(p_t \Vert q_t\big)
:= \int \log r\,\mathcal{L}p\,d\mathbf{x} - \int r\,\mathcal{L}q\,d\mathbf{x}.
\]

\textbf{Step 2: Drift does not change the KL.} For the drift operator $-\nabla\cdot(fu)$, integration by parts (with vanishing boundary terms) gives
\[
\int \log r\,\big[-\nabla\cdot(fp)\big]\,d\mathbf{x}
 = \int p\,f\cdot\nabla\log r\,d\mathbf{x},
\]
\[
\int r\,\big[-\nabla\cdot(fq)\big]\,d\mathbf{x}
 = \int q\,f\cdot\nabla r\,d\mathbf{x}.
\]

Using $r = p/q$ and $\nabla\log r = \nabla r / r$, one checks
\[
p\,f\cdot\nabla\log r - q\,f\cdot\nabla r = 0,
\]

so the \textbf{drift part cancels exactly} and does not affect $\mathrm{KL}(p_t\Vert q_t)$.

\textbf{Step 3: Diffusion decreases the KL.} For the diffusion operator $\tfrac{1}{2}g(t)^2\nabla^2 u$, integration by parts yields
\[
\int \log r\cdot\frac{1}{2}g(t)^2\nabla^2 p\,d\mathbf{x}
 = -\frac{1}{2}g(t)^2\int \nabla\log r\cdot\nabla p\,d\mathbf{x},
\]
\[
\int r\cdot\frac{1}{2}g(t)^2\nabla^2 q\,d\mathbf{x}
 = -\frac{1}{2}g(t)^2\int \nabla r\cdot\nabla q\,d\mathbf{x}.
\]

Using
\[
\nabla p = p\,\nabla\log p, \qquad
\nabla q = q\,\nabla\log q,\qquad
\nabla r = \nabla\!\left(\frac{p}{q}\right)
         = r\big(\nabla\log p - \nabla\log q\big),
\]

we obtain
\[
\nabla\log r\cdot\nabla p
:= p\big(\nabla\log p - \nabla\log q\big)\cdot\nabla\log p,
\]
\[
\nabla r\cdot\nabla q
:= p\big(\nabla\log p - \nabla\log q\big)\cdot\nabla\log q.
\]

Subtracting these contributions gives
\[
-\frac{1}{2}g(t)^2\int \nabla\log r\cdot\nabla p\,d\mathbf{x}
+\frac{1}{2}g(t)^2\int \nabla r\cdot\nabla q\,d\mathbf{x}
:= -\frac{1}{2}g(t)^2\int p(\mathbf{x},t)
   \big\|\nabla\log p - \nabla\log q\big\|^2\,d\mathbf{x}.
\]

\textbf{Step 4: Conclusion.} Putting drift and diffusion together,
\[
\frac{d}{dt}\mathrm{KL}\big(p_t\Vert q_t\big)
:= -\frac{1}{2}g(t)^2
   \int p(\mathbf{x},t)\,
        \big\|\nabla\log p(\mathbf{x},t)
              - \nabla\log q(\mathbf{x},t)\big\|^2
      d\mathbf{x}
\;\le 0.
\]

Thus, along the forward diffusion process, the KL divergence between any two solutions of the same Fokker--Planck equation is \textbf{non-increasing}: diffusion strictly contracts KL (with equality only if the scores $\nabla\log p$ and $\nabla\log q$ coincide almost everywhere). This monotone decrease of $\mathrm{KL}(p_t \Vert q_t)$ justifies decomposing the global maximum-likelihood objective into local-in-time, squared-score terms associated with each diffusion step.

\subsection{Derivation: equivalence between DSM and SM}
\label{sec:appendix-dsm-sm}

We now prove that the \textit{denoising score matching} (DSM) loss and the \textit{score matching} (SM) loss at time $t$ have the same minimizer.

\textbf{Step 1: Define the two losses.} Let us write the \textit{denoising score matching} (DSM) loss at time $t$ as
\[
L_{\text{DSM}}(\mathbf{s}_\theta)
:= \mathbb{E}_{\mathbf{x}_0 \sim p_0}\,
   \mathbb{E}_{\mathbf{x}_t \sim p_t(\cdot \mid \mathbf{x}_0)}
   \big\|
      \nabla_{\mathbf{x}_t} \log p_t(\mathbf{x}_t \mid \mathbf{x}_0)
      - \mathbf{s}_\theta(\mathbf{x}_t, t)
   \big\|^2,
\]
and the \textit{score matching} (SM) loss on the marginal $p_t(\mathbf{x}_t)$ as
\[
L_{\text{SM}}(\mathbf{s}_\theta)
:= \mathbb{E}_{\mathbf{x}_t \sim p_t}
   \big\|
      \nabla_{\mathbf{x}_t} \log p_t(\mathbf{x}_t)
      - \mathbf{s}_\theta(\mathbf{x}_t, t)
   \big\|^2.
\]

Here $p_t(\mathbf{x}_t) = \int p_t(\mathbf{x}_t \mid \mathbf{x}_0)\,p_0(\mathbf{x}_0)\,d\mathbf{x}_0$ is the marginal of the forward process at time $t$.

\textbf{Step 2: Introduce conditional and marginal scores.} Define the conditional score
\[
\mathbf{s}(\mathbf{x}_t \mid \mathbf{x}_0)
:= \nabla_{\mathbf{x}_t} \log p_t(\mathbf{x}_t \mid \mathbf{x}_0),
\]
and the marginal score
\[
\mathbf{s}(\mathbf{x}_t, t)
:= \nabla_{\mathbf{x}_t} \log p_t(\mathbf{x}_t).
\]

\textbf{Step 3: Expand both objectives.} Using $\|\mathbf{a}-\mathbf{b}\|^2 = \|\mathbf{a}\|^2 + \|\mathbf{b}\|^2 - 2\langle \mathbf{a}, \mathbf{b}\rangle$, we can expand both objectives. For DSM,
\[
\begin{aligned}
L_{\text{DSM}}(\mathbf{s}_\theta)
&= \mathbb{E}_{\mathbf{x}_0, \mathbf{x}_t}
    \big\| \mathbf{s}_\theta(\mathbf{x}_t, t) \big\|^2
   - 2\,\mathbb{E}_{\mathbf{x}_0, \mathbf{x}_t}
       \big\langle
          \mathbf{s}_\theta(\mathbf{x}_t, t),
          \mathbf{s}(\mathbf{x}_t \mid \mathbf{x}_0)
       \big\rangle \\
&\quad
   + \mathbb{E}_{\mathbf{x}_0, \mathbf{x}_t}
       \big\|\mathbf{s}(\mathbf{x}_t \mid \mathbf{x}_0)\big\|^2,
\end{aligned}
\]

where expectations are taken under the joint $p_0(\mathbf{x}_0)\,p_t(\mathbf{x}_t \mid \mathbf{x}_0)$. Similarly, for SM we have
\[
\begin{aligned}
L_{\text{SM}}(\mathbf{s}_\theta)
&= \mathbb{E}_{\mathbf{x}_t}
    \big\| \mathbf{s}_\theta(\mathbf{x}_t, t) \big\|^2
   - 2\,\mathbb{E}_{\mathbf{x}_t}
       \big\langle
          \mathbf{s}_\theta(\mathbf{x}_t, t),
          \mathbf{s}(\mathbf{x}_t, t)
       \big\rangle \\
&\quad
   + \mathbb{E}_{\mathbf{x}_t}
       \big\|\mathbf{s}(\mathbf{x}_t, t)\big\|^2.
\end{aligned}
\]

\textbf{Step 4: Match the first and last terms.} The first terms coincide, because the marginal of the joint distribution is exactly $p_t(\mathbf{x}_t)$:
\[
\mathbb{E}_{\mathbf{x}_0, \mathbf{x}_t}
  \big\|\mathbf{s}_\theta(\mathbf{x}_t, t)\big\|^2
 = \int p_t(\mathbf{x}_t)
       \big\|\mathbf{s}_\theta(\mathbf{x}_t, t)\big\|^2
   \,d\mathbf{x}_t
 = \mathbb{E}_{\mathbf{x}_t}
     \big\|\mathbf{s}_\theta(\mathbf{x}_t, t)\big\|^2.
\]

The last terms,
\[
\mathbb{E}_{\mathbf{x}_0, \mathbf{x}_t}\|\mathbf{s}(\mathbf{x}_t \mid \mathbf{x}_0)\|^2
\]
and
\[
\mathbb{E}_{\mathbf{x}_t}\|\mathbf{s}(\mathbf{x}_t, t)\|^2,
\]
do \textbf{not} depend on $\mathbf{s}_\theta$ at all, so they can only shift the loss by a constant.

\textbf{Step 5: Handle the cross term.} The only subtle point is the cross term. Because the inner product is linear, it is enough to prove that, for any (scalar) test function $f(\mathbf{x}_t)$,
\[
\mathbb{E}_{\mathbf{x}_0, \mathbf{x}_t}
  \big[ f(\mathbf{x}_t)\,\mathbf{s}(\mathbf{x}_t \mid \mathbf{x}_0) \big]
=
\mathbb{E}_{\mathbf{x}_t}
  \big[ f(\mathbf{x}_t)\,\mathbf{s}(\mathbf{x}_t, t) \big],
\]
and then apply this to each coordinate of $\mathbf{s}_\theta(\mathbf{x}_t, t)$.

By definition of the score,
\[
\mathbf{s}(\mathbf{x}_t \mid \mathbf{x}_0)
:= \nabla_{\mathbf{x}_t} \log p_t(\mathbf{x}_t \mid \mathbf{x}_0)
:= \frac{\nabla_{\mathbf{x}_t} p_t(\mathbf{x}_t \mid \mathbf{x}_0)}
       {p_t(\mathbf{x}_t \mid \mathbf{x}_0)}.
\]

Therefore,
\[
\begin{aligned}
\mathbb{E}_{\mathbf{x}_0, \mathbf{x}_t}
  \big[ f(\mathbf{x}_t)\,\mathbf{s}(\mathbf{x}_t \mid \mathbf{x}_0) \big]
&= \iint
     p_0(\mathbf{x}_0)\,
     p_t(\mathbf{x}_t \mid \mathbf{x}_0)\,
     f(\mathbf{x}_t)\,
     \frac{\nabla_{\mathbf{x}_t} p_t(\mathbf{x}_t \mid \mathbf{x}_0)}
          {p_t(\mathbf{x}_t \mid \mathbf{x}_0)}
     \,d\mathbf{x}_t\,d\mathbf{x}_0 \\
&= \iint
     f(\mathbf{x}_t)\,
     \nabla_{\mathbf{x}_t} p_t(\mathbf{x}_t \mid \mathbf{x}_0)\,
     p_0(\mathbf{x}_0)\,
     \,d\mathbf{x}_t\,d\mathbf{x}_0.
\end{aligned}
\]

Under mild regularity conditions we can interchange the order of integration and differentiation, obtaining
\[
\begin{aligned}
\mathbb{E}_{\mathbf{x}_0, \mathbf{x}_t}
  \big[ f(\mathbf{x}_t)\,\mathbf{s}(\mathbf{x}_t \mid \mathbf{x}_0) \big]
&= \int
     f(\mathbf{x}_t)\,
     \nabla_{\mathbf{x}_t}
       \Big(
         \int p_t(\mathbf{x}_t \mid \mathbf{x}_0)\,p_0(\mathbf{x}_0)\,d\mathbf{x}_0
       \Big)
     \,d\mathbf{x}_t \\
&= \int f(\mathbf{x}_t)\,\nabla_{\mathbf{x}_t} p_t(\mathbf{x}_t)\,d\mathbf{x}_t \\
&= \int
     p_t(\mathbf{x}_t)\,
     f(\mathbf{x}_t)\,
     \nabla_{\mathbf{x}_t} \log p_t(\mathbf{x}_t)\,d\mathbf{x}_t \\
&= \mathbb{E}_{\mathbf{x}_t}
     \big[ f(\mathbf{x}_t)\,\mathbf{s}(\mathbf{x}_t, t) \big].
\end{aligned}
\]

Taking $f(\mathbf{x}_t)$ to be each component of $\mathbf{s}_\theta(\mathbf{x}_t, t)$ shows that the DSM and SM cross terms are identical:
\[
\mathbb{E}_{\mathbf{x}_0, \mathbf{x}_t}
  \big\langle
     \mathbf{s}_\theta(\mathbf{x}_t, t),
     \mathbf{s}(\mathbf{x}_t \mid \mathbf{x}_0)
  \big\rangle
=
\mathbb{E}_{\mathbf{x}_t}
  \big\langle
     \mathbf{s}_\theta(\mathbf{x}_t, t),
     \mathbf{s}(\mathbf{x}_t, t)
  \big\rangle.
\]

\textbf{Conclusion.} Putting everything together, we have
\[
L_{\text{DSM}}(\mathbf{s}_\theta)
:= L_{\text{SM}}(\mathbf{s}_\theta) + C,
\]
where $C$ is a constant independent of $\mathbf{s}_\theta$. Hence both objectives are minimized by the same function, namely the true marginal score
\[
\mathbf{s}_\theta^\star(\mathbf{x}_t, t) = \nabla_{\mathbf{x}_t} \log p_t(\mathbf{x}_t).
\]

\bibliographystyle{plainnat}
\bibliography{references}

@article{Luo2022UnderstandingDM,
  title={Understanding Diffusion Models: A Unified Perspective},
  author={Luo, Calvin},
  journal={arXiv preprint arXiv:2208.11970},
  year={2022},
  url={https://arxiv.org/abs/2208.11970}
}

@article{Ho2020DenoisingDP,
  title={Denoising Diffusion Probabilistic Models},
  author={Ho, Jonathan and Jain, Ajay and Abbeel, Pieter},
  journal={arXiv preprint arXiv:2006.11239},
  year={2020},
  url={https://arxiv.org/abs/2006.11239}
}

@article{Song2020ScoreBasedGM,
  title={Score-Based Generative Modeling through Stochastic Differential Equations},
  author={Song, Yang and Sohl-Dickstein, Jascha and Kingma, Diederik P. and Kumar, Abhishek and Ermon, Stefano and Poole, Ben},
  journal={arXiv preprint arXiv:2011.13456},
  year={2020},
  url={https://arxiv.org/abs/2011.13456}
}

@article{song2019generative,
  title={Generative modeling by estimating gradients of the data distribution},
  author={Song, Yang and Ermon, Stefano},
  journal={Advances in Neural Information Processing Systems},
  year={2019},
  url={https://arxiv.org/abs/1907.05600}
}

@article{Anderson1982ReversetimeDE,
  title={Reverse-time diffusion equation models},
  author={Anderson, Brian D. O.},
  journal={Stochastic Processes and their Applications},
  year={1982},
  url={https://doi.org/10.1016/0304-4149(82)90051-5}
}

@article{liu2022flow,
  title={Flow straight and fast: Learning to generate and transfer data with rectified flow},
  author={Liu, Xingchao and Gong, Chengyue and Liu, Qiang},
  journal={arXiv preprint arXiv:2209.03003},
  year={2022},
  url={https://arxiv.org/abs/2209.03003}
}

@article{Karras2022Elucidating,
  title={Elucidating the Design Space of Diffusion-Based Generative Models},
  author={Karras, Tero and Aittala, Miika and Aila, Timo and Laine, Samuli},
  journal={arXiv preprint arXiv:2206.00364},
  year={2022},
  url={https://arxiv.org/abs/2206.00364}
}

@inproceedings{gao2025diffusion,
  title={Diffusion Models and Gaussian Flow Matching: Two Sides of the Same Coin},
  author={Gao, Ruiqi and Hoogeboom, Emiel and Heek, Jonathan and De Bortoli, Valentin and Murphy, Kevin Patrick and Salimans, Tim},
  booktitle={The Fourth Blogpost Track at ICLR 2025},
  year={2025},
  url={https://openreview.net/forum?id=C8Yyg9wy0s}
}

@article{Langevin1908,
  title={Sur la th{\'e}orie du mouvement brownien},
  author={Langevin, Paul},
  journal={Comptes Rendus de l'Acad{\'e}mie des Sciences},
  volume={146},
  pages={530--533},
  year={1908}
}

@article{zheng2025lanpaint,
  title={LanPaint: Training-Free Diffusion Inpainting with Asymptotically Exact and Fast Conditional Sampling},
  author={Zheng, Candi and Lan, Yuan and Wang, Yang},
  journal={Transactions on Machine Learning Research},
  year={2025},
  url={https://openreview.net/forum?id=JPC8JyOUSW}
}

\end{document}